\documentclass[10pt,twocolumn,letterpaper]{article}
\usepackage{multirow}
\usepackage{pifont}

\usepackage{booktabs}
\usepackage{todonotes}
\usepackage{amsmath}
\usepackage{bm}
\usepackage{graphics}
\usepackage{makecell}
\usepackage[accsupp]{axessibility}

\DeclareMathOperator*{\argmax}{arg\,max}

\definecolor{mygreen}{RGB}{0, 163, 108}

\newcommand{\myparagraph}[1]{\vspace{0.2cm}\noindent\textbf{#1}.\ }

\newcommand{\method}{{Prototype-based Masked Cross-Attention}}
\newcommand{\methodSpace}{{Prototype-based Masked Cross-Attention} }
\newcommand{\architecture}{{PEM}}
\newcommand{\architectureSpace}{{PEM} }
\newcommand*\samethanks[1][\value{footnote}]{\footnotemark[#1]}

\usepackage[pagenumbers]{cvpr} 

\definecolor{cvprblue}{rgb}{0.21,0.49,0.74}
\usepackage[pagebackref,breaklinks,colorlinks,citecolor=cvprblue]{hyperref}

\def\paperID{2492} 
\def\confName{CVPR}
\def\confYear{2024}

\title{PEM: Prototype-based Efficient MaskFormer for Image Segmentation}

\author{Niccolò Cavagnero\thanks{Equal contribution}\ $^{,1}$, \hspace{2mm}Gabriele Rosi\samethanks\ $^{,1,2}$, \hspace{2mm}Claudia Cuttano$^1$, \hspace{2mm}Francesca Pistilli$^1$, \\\hspace{2mm}Marco Ciccone$^1$, Giuseppe Averta$^{1,2}$, \hspace{2mm}Fabio Cermelli$^2$ \\
$^1$ Politecnico di Torino, $^2$ Focoos AI\\
$^1$ {\tt\small name.surname@polito.it}, $^2$ {\tt\small name.surname@focoos.ai}
}

\begin{document}
\maketitle 
\begin{abstract}
Recent transformer-based architectures have shown impressive results in the field of image segmentation. Thanks to their flexibility, they obtain outstanding performance in multiple segmentation tasks, such as semantic and panoptic, under a single unified framework. 
To achieve such impressive performance, these architectures employ intensive operations and require substantial computational resources, which are often not available, especially on edge devices.
To fill this gap, we propose Prototype-based Efficient MaskFormer (PEM), an efficient transformer-based architecture that can operate in multiple segmentation tasks. PEM proposes a novel prototype-based cross-attention which leverages the redundancy of visual features to restrict the computation and improve the efficiency without harming the performance. 
In addition, PEM introduces an efficient multi-scale feature pyramid network, capable of extracting features that have high semantic content in an efficient way, thanks to the combination of deformable convolutions and context-based self-modulation.
We benchmark the proposed PEM architecture on two tasks, semantic and panoptic segmentation, evaluated on two different datasets, Cityscapes and ADE20K. PEM demonstrates outstanding performance on every task and dataset, outperforming task-specific architectures while being comparable and even better than computationally expensive baselines. Code is available at \href{https://github.com/NiccoloCavagnero/PEM}{https://github.com/NiccoloCavagnero/PEM}.
\end{abstract}

\begin{figure}[t]
    \centering
    \includegraphics[width=0.99\linewidth]{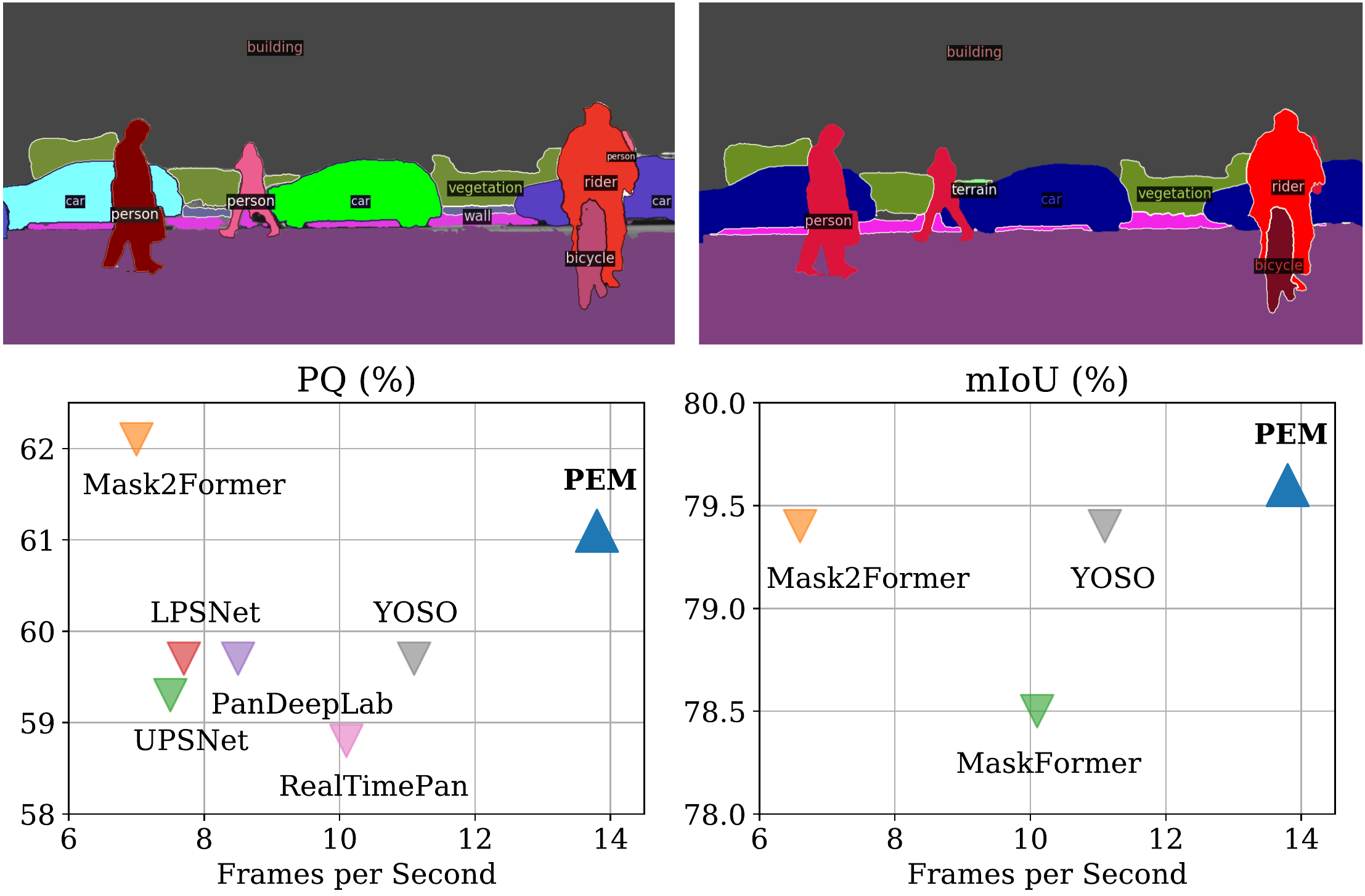}
    \caption{PEM delivers comparable or superior performance in comparison to existing methods while being the fastest multi-task architecture for image segmentation.} \vspace{-11pt}
    \label{fig:teaser}
\end{figure}

\vspace{-0.5cm}
\section{Introduction}
\label{intro}
Image segmentation stands as a cornerstone within the realm of computer vision and image processing, playing a pivotal role in the extraction of meaningful information from digital images. At its core, it involves partitioning an image into distinct regions, or segments, each representing a significant object or component within the visual scene.

One of the recent and noteworthy developments in this domain is the emergence of transformer-based approaches \cite{maskformer, m2f, kmax, kirillov2023segment}, which offer a unified framework for the various flavors of image segmentation, such as semantic and panoptic segmentation. These architectures rely on an end-to-end set prediction objective, inspired by DETR \cite{detr}, and they employ an encoder-decoder architecture to generate high-resolution visual features and a transformer decoder to yield a representation for each object. By leveraging the same architecture, loss function and training pipeline for different segmentation tasks, these methods obtain outstanding results, outperforming task-specific architectures while streamlining the segmentation process as a whole. 

To achieve their remarkable performance and appealing properties, however, these models require expensive architectural components, resulting in slow and cumbersome architectures. The overly inefficient inference of existing methods leads to two important consequences: i) the deployment and inference cost, as well as the carbon footprint \cite{patterson2021carbon, gowda2023watt}, is not negligible, especially when offering cloud services to millions of users; ii) the computational demand makes the deployment on edge devices unfeasible, preventing the use on resource-constrained downstream tasks. 

In this paper, we rethink their approach with the goal of doing more with less. We propose a novel image segmentation architecture, named \textbf{P}rototype-based \textbf{E}fficient \textbf{M}askFormer (PEM), that properly considers the complexity of each component and it substitutes standard computationally demanding choices with novel and lighter counterparts.
We first focus on the transformer decoder, originally constituted by a sequence of expensive attention operations between object descriptors and high-resolution image features. To reduce the computation and enhance scalability with increasing image resolutions, we incorporate a prototype selection mechanism which leverages a single visual token for each object descriptor. Furthermore, the introduction of prototypes allows us to design a novel efficient attention mechanism. 
Second, we revisit the visual decoder which is crucial for the extraction of high-resolution features. While previous works enhanced a feature pyramid network (FPN) with transformer-based attention modules \cite{maskformer, m2f, kmax}, we employ a more efficient fully convolutional FPN. We supplement it with a context-based self-modulation module to recover the global context and deformable convolutions \cite{deformable_conv} to allow each kernel to dynamically modify its receptive field to focus on relevant regions.

We benchmark the proposed PEM architecture on two distinct tasks, semantic and panoptic segmentation, on two datasets: Cityscapes \cite{cityscapes} and ADE20K \cite{ade} (see \cref{fig:teaser}). PEM exhibits outstanding performance, showcasing similar or superior results compared to the computationally expensive baselines. Remarkably, PEM is able to outperform task-specific architectures on the challenging ADE20K. 

By addressing both the major bottlenecks of modern segmentation models, PEM represents a significant step forward in the ongoing pursuit of efficient image segmentation methodologies, making them more sustainable and amenable to real-world applications. 
In summary, this paper provides the following contributions:
\begin{itemize}
    \item A novel efficient cross-attention mechanism endowed with a prototype selection strategy that lowers the computational burden without affecting the results; 
    \item A multi-scale FPN that presents the benefits of heavy transformer-based decoders in a convolutional fashion; 
    \item Through comprehensive quantitative and qualitative analysis, we showcase the generality, efficiency, and state-of-the-art performance of PEM on both semantic and panoptic segmentation on two challenging benchmarks. 
\end{itemize}

\begin{figure*}[t]
\vspace{-1cm}
    \centering
    \includegraphics[width=1\linewidth]{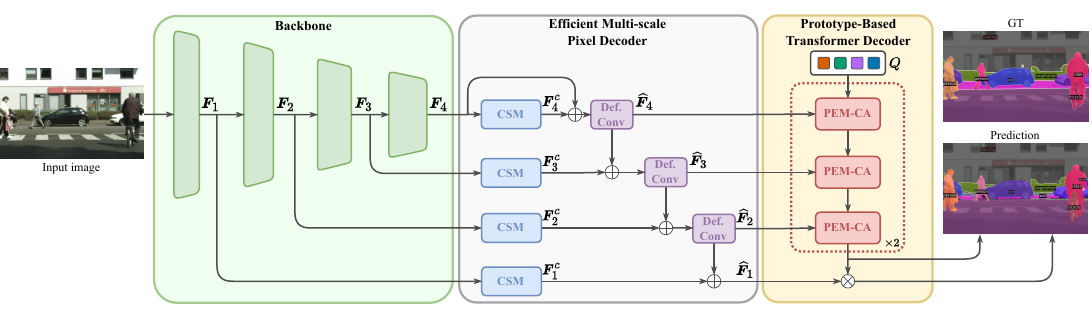}
    \caption{\textbf{Architecture of \architecture} with the three main components highlighted: backbone, pixel decoder and transformer decoder. The backbone extracts features from the input image; the pixel decoder provides features upsampling to extract high-resolution features; the transformer decoder, which takes as input a set of learnable queries and the high-resolution features to produce refined queries for inference.} \vspace{-5pt}
    \label{fig:architecture}
\end{figure*}

\section{Related Works}
\label{related}

\myparagraph{Image Segmentation}
A growing field of research aspires to design architecture able to operate in multiple image segmentation settings, without any change in loss function or architectural component. The seminal work of DETR \cite{detr} showed that it is possible to achieve competitive object detection and panoptic segmentation results using an end-to-end set prediction network based on mask classification. 
Inspired by this work, MaskFormer \cite{maskformer} proposed an architecture for image segmentation based on the mask classification approach, achieving state-of-the-art performance both in semantic and in panoptic segmentation. Mask2Former \cite{m2f} further improved it, proposing various architectural enhancements that led to faster convergence and results that outperformed not only general-purpose approaches but also specialized architectures.  
More recently, kMaX-DeepLab \cite{kmax} attempted to improve the cross-attention mechanism by replacing the classic attention with k-Means clustering operation. 
While significant progress has been made in improving overall performance across all tasks, the high resource requirements and slow inference time still hinder the deployment of these models on edge devices.

\myparagraph{Efficient Image Segmentation}
To reduce the computational complexity of image segmentation models, multiple works \cite{bisenet, stdc, pidnet, wang2020solo, pandeeplab, realtimepan, yoso} proposed efficient network architectures which can be effectively deployed on devices and run in real-time. However, previous works proposed architectures specific for a single segmentation task only. 
Efficient semantic segmentation works were based on a two-branch architecture \cite{bisenet, bisenetv2, stdc, ddrnet} in which high-resolution and highly semantic features are separately processed and then merged. Recently, PIDNet \cite{pidnet} proposed a three-branch semantic segmentation architecture to enhance the object boundaries and improve performance on small objects. 
In panoptic segmentation, UPSNet \cite{upsnet} proposed a network that incorporates a parameter-free panoptic head to efficiently merge the predictions of segmentation and instance heads. Meanwhile, FPSNet \cite{fpsnet} eliminated the need for additional segmentation heads and it introduced an architecture based on attention mask merging. Other approaches \cite{lpsnet,pandeeplab,realtimepan} localized objects using points or boxes. Recently, YOSO \cite{yoso} proposed an efficient method that predicts masks for both things and stuff using a transformer architecture. 
Despite significant progress achieved in various settings, these architectures only operate on a single task, duplicating the research efforts and jeopardizing the research landscape. Our work aims to address this gap by presenting an efficient architecture that can be seamlessly employed in multiple segmentation tasks.

\myparagraph{Efficient Attention Mechanism}
Since its introduction, Attention \cite{attention} has exhibited remarkable capabilities in several tasks thanks to its generality and its ability to model global relationships among all input elements. MaskFormer and its variants~\cite{maskformer, m2f, kmax} established both self-attention and cross-attention as two fundamental components in their architectures. Nevertheless, by considering all pairwise relationships among input tokens, the attention mechanism does not scale effectively for large input dimensions. This fundamental issue has prompted the scientific community to explore methods for mitigating the computational complexity of this module.
Some approaches aimed to reduce the computational cost of self-attention by integrating downsampling operators, typically pooling layers or strided convolutions, within the projections \cite{efv2}. Consequently, the self-attention operation is performed at a lower resolution, improving efficiency.
MobileViT V2 \cite{separable} and SwiftFormer \cite{swift} took a step forward, by replacing the expensive dot products between the input tokens with cheap element-wise multiplications of the features with a small context vector. These methods demonstrated how pairwise interactions can be redundant and global information can be condensed into lightweight context vectors, which are computationally inexpensive.
While these works focus on reducing the complexity of self-attention, we propose a method for efficiently computing cross-attention, which constitutes one of the major bottlenecks in MaskFormer architectures. Our custom cross-attention effectively combines visual tokens and object queries to yield fast and precise image segmentation.

\section{Method}
\label{method}

\subsection{Mask-based Segmentation Framework} 
In the field of image segmentation, the objective is to partition images into regions that exhibit shared characteristics. This includes tasks as semantic segmentation, where pixels with similar semantic attributes are grouped together, and panoptic segmentation, which involves differentiating instances within the same class. The goal is to develop a model that segments the image into separate regions with unique masks with assigned class probabilities.
Formally, given an image, $\bm{I} \in \mathbb{R}^{H \times W \times 3}$, we aim to predict a set of $N$ binary masks $\bm{M} \in \{0, 1\}^{N \times H \times W}$ each associated with a probability distribution $\bm{p}_i \in \Delta^{K+1}$, where $(H, W)$ is the height and width of the image, and $K+1$ is the number of classes plus an additional ``\textit{no object}'' class.

To achieve this goal, we follow the framework provided by the MaskFormer architecture~\cite{maskformer, m2f, kmax}  that consists of three main components, depicted in \cref{fig:architecture}: (i) a \textit{backbone} extracting feature maps $\bm{F}_i \in \mathbb{R}^{H_i \times W_i \times B_i}$ from the image $\bm{I}$, (ii) \textit{a pixel decoder} that processes $\bm{F}_i$ to produce high-resolution multi-scale features $\widehat{\bm{F}}_i \in \mathbb{R}^{H_i \times W_i \times C}$, with $i \in \{1,2,3,4\}$ and $H_i$, $W_i$ equal to the image resolution divided by, respectively, $4$, $8$, $16$, and $32$, and (iii) a \textit{transformer decoder} which accepts three multi-scale features $\widehat{\bm{F}}_i$, $i \in \{2,3,4\}$ as input together with $N$ learnable queries $\bm{Q} \in \mathbb{R}^{N \times C}$ and it generates $N$ refined queries $\widehat{\bm{Q}} \in \mathbb{R}^{N \times C}$. 
To generate the $N$ binary masks, the refined queries $\widehat{\bm{Q}}$ are then multiplied by the highest resolution features of the pixel decoder $\widehat{\bm{F}}_1$. Finally, the class probabilities are obtained through a linear classifier applied on $\widehat{\bm{Q}}$.

\begin{figure*}[t]
    \vspace{-1cm}
    \centering
        \includegraphics[width=.82\textwidth]{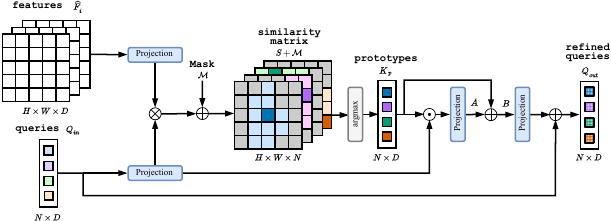}
    \caption{\textbf{Scheme of the proposed \method.} The prototype selection mechanism reduces the token dimension from HW to N, the number of queries, significantly reducing the computational burden.
    } \vspace{-10pt}
    \label{fig:protoCA}
\end{figure*}

\subsection{Prototype-based Masked Cross-Attention}
A core component of MaskFormer architecture family \cite{maskformer, m2f, kmax} is the transformer decoder which, taking as input $N$ learnable queries and high-resolution image features, has the objective of refining the queries which are later used to obtain the predictions. This is achieved through several transformer blocks that attend to the feature representations and model relations between different objects. Each block computes a cross-attention between visual features and the object queries, acting as anchors, relying on expensive dot products.
Despite its remarkable performance, it is inevitably inefficient when applied to large input features, which are typical in segmentation tasks. 

In this work, we improve the efficiency of this module by proposing an architectural enhancement, denoted as \methodSpace (PEM-CA) and illustrated in \cref{fig:protoCA}. First, PEM-CA capitalizes on the intrinsic redundancy of visual features in segmentation to significantly reduce the number of input tokens in attention layers through a prototype selection mechanism. Indeed, during training, features related to the same segment naturally align and we can therefore exploit this redundancy to process only a subset of the visual tokens. Second, inspired by recent advancements in the efficiency of attention modules~\cite{mobilevit, swift}, PEM-CA redesigns the cross-attention operation, modeling interactions by means of computationally cheap element-wise operations.

\myparagraph{Prototype Selection} \label{sec:pemca}
The goal of the cross-attention is to refine each input query based on the visual features of the object it represents. However, we argue that using all the pixels belonging to an object for the refinement is redundant, since pixels associated with a specific object query will naturally become close to each other as training progresses.
We can leverage this inherent redundancy to focus solely on the most relevant feature for each object, \ie \textit{the prototype}, while discarding the others for subsequent operations and reducing the computation.

In practice, to compute the prototypes, we first project the high-resolution features $\widehat{\bm{F}}_i \in \mathbb{R}^{H_i \times W_i \times C}$ and the object queries $\bm{Q}_{in} \in \mathbb{R}^{N \times C}$ in the same dimensional space and we obtain, respectively, $\bm{K} \in \mathbb{R}^{H_i W_i \times D}$ and $\bm{Q} \in \mathbb{R}^{N \times D}$. 
Then, the similarity matrix $\bm{S} \in \mathbb{R}^{H_i W_i \times N}$ is computed as: 
\begin{equation}
    \bm{S} = \bm{K} \bm{Q}^T, 
\end{equation}
which represents the relationship between each pixel of the image feature with the object queries.
Once $\bm{S}$ is obtained, for each query, we select its most similar pixel according to $\textbf{S}$. The subset of selected pixels is termed \textit{prototypes}, as they represent the most representative token of their respective object.
Formally, the prototypes $\bm{K}_{p}$ are computed as:
\begin{align}
    \centering
    \bm{G} &= \argmax_{H_i W_i}(\bm{S} + \bm{\mathcal{M}} ),\\
    \bm{K}_{p} &= \bm{K} \left[\bm{G} \right],
    \label{prototype_selection}
\end{align}
where $\bm{\mathcal{M}}$ is a binary mask applied to the similarity matrix, $\bm{G}$ the selected token indices and $\bm{K} \left[\bm{G}\right]$ denotes the selection of the indices in $\bm{G}$ on the first dimension of $\bm{K}$.
The binary mask $\bm{\mathcal{M}}$ introduces a masking mechanism as in \cite{m2f} that forces the selection to only consider foreground pixels, leading to more consistent assignments and improved training performance. We compute $\bm{\mathcal{M}}$ as:
\begin{align}
    \centering
    \bm{\mathcal{M}}(x,y) &=
        \begin{cases}
            0 & \text{if } \bm{M}_{l-1}(x,y) = 1 \\
            -\infty & \text{otherwise},
        \end{cases}
\end{align}
where $\bm{M}$ is the binarized output of the previous transformer decoder layer resized at the same resolution of $\widehat{\bm{F}}_i$.

We note that the whole selection process is performed in a multi-head fashion, with features and queries divided in heads across the channel dimension. Hence, each group of channels of a token can be assigned to the corresponding group of a given query, thereby improving the modeling capability of the selection process.

\begin{figure}[t]
    \centering 
    \includegraphics[width=0.99\linewidth]{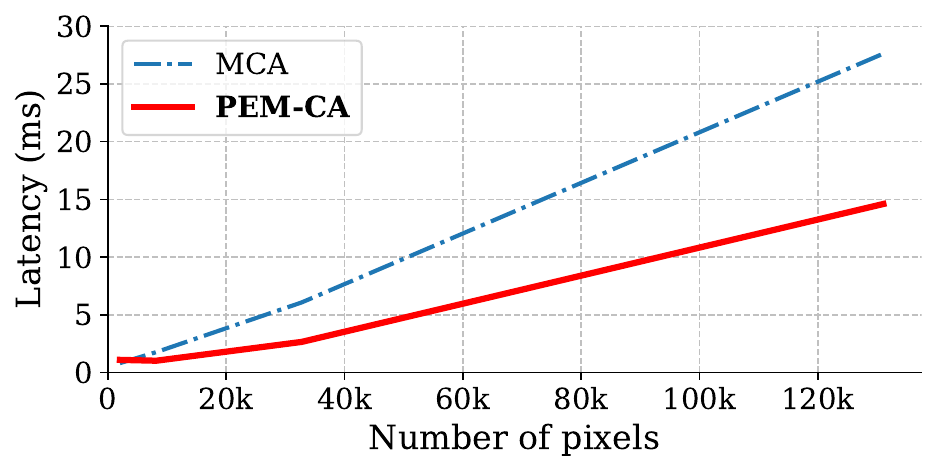}
    \caption{\textbf{Latency comparison between PEM-CA and Masked Cross-Attention.} PEM-CA scales better \wrt Masked Cross-Attention \cite{m2f} when the input dimension increases. Note that, for Cityscapes images (1024$\times$2048 pixels), the features have dimensions 2048 ($F_4$), 8192 ($F_3$), 32768 ($F_2$), 131072 ($F_1$) {pixels}.} \vspace{-12pt}
    \label{fig:latency}
\end{figure}

\myparagraph{Prototype-based Cross-Attention}
The prototype selection mechanism reduces the input from $\bm{K} \in \mathbb{R}^{HW \times D}$ to $\bm{K}_{p} \in \mathbb{R}^{N \times D}$, significantly decreasing the complexity of any subsequent operation. In addition, it establishes a matching between the queries and their corresponding prototype. 
Consequently, the $\bm{Q}$-$\bm{K}_{p}$ interaction can be modeled through a cheap element-wise product and a projection $\bm{W}_A \in \mathbb{R}^{D \times D}$, avoiding the need of leveraging all pairwise relationships. Formally,
\begin{equation}
\bm{A} = (\bm{Q} \odot \bm{K}_{p})\bm{W}_A.
\end{equation}
The matrix $\bm{A}$ is then normalized across the channel dimension and scaled by a learnable parameter $\boldsymbol{\alpha}\in\mathbb{R}^{D}$.
The scaled attention matrix indicates the strength of interaction between prototypes and queries, and we use it to dynamically reweight the prototypes $\bm{K}_{p}$ in an additive manner:
\begin{equation}
          \bm{B} =\boldsymbol{\alpha} \odot \frac{\bm{A}}{||\bm{A}||_2}  + \bm{K}_{p}.
\end{equation}
To obtain the output $\bm{Q}_{out}$, a final linear projection $\bm{W}_{out} \in \mathbb{R}^{D \times C}$ brings the hidden states of the module back to the queries space before applying a residual connection:
\begin{align}
      \bm{Q}_{out} = \bm{B} \bm{W}_{out} + \bm{Q}.
\end{align}
The incorporation of PEM-CA allows a more efficient interaction between visual features and object queries compared to traditional cross-attention mechanisms. As depicted in Figure \ref{fig:latency}, it is evident that as the resolution of the input features increases, the gap in terms of latency also widens. PEM-CA demonstrates a notable advantage, being $2\times$ faster than the masked cross-attention counterpart. This observation underscores the efficiency gains achieved through the adoption of PEM-CA in managing the computational demands of attention mechanisms.

\subsection{Efficient Multi-scale Pixel Decoder}
The pixel decoder covers a fundamental role in extracting multi-scale features which allow a precise segmentation of the objects. Mask2Former \cite{m2f} implements it as a feature pyramid network (FPN) enhanced with deformable attention. Deformable attention is characterized by three fundamental properties: (i) it attends global context for feature refinement, (ii) it computes dynamic weights based on the input, and (iii) it leverages deformability to make the receptive field input-dependent, favoring the focus on relevant regions for the input. 
Despite its performance gains, using deformable attention upon an FPN introduces a computation overhead that makes the pixel decoder inefficient and unsuitable for real-world applications. 
To maintain the performance while being computationally efficient, we use a fully convolutional FPN where we restore the benefits of deformable attention by leveraging two key techniques. 
First, to reintroduce the global context (i) and the dynamic weights (ii), we implement context-based self-modulation (CSM) modules that adjust the input channels using a global scene representation \cite{se}. Moreover, to enable deformability (iii), we adopt deformable convolutions that focus on relevant regions of the image by dynamically adapt the receptive field. This dual approach yields competitive performance while preserving the computational efficiency. 

\myparagraph{Context-based Self-Modulation}
Features coming from the backbone are highly localized and contain rich spatial details but they lack a general understanding of the scene context. 
To restore and efficiently inject this information at all scales, we take inspiration from previous works \cite{perez2018film, se, bisenet, stdc} and employ a context-based self-modulation (CSM) mechanism to reweight the importance of each channel based on a global scene representation, which suppresses less informative channels and enhances the most informative ones. 
Specifically, given the input features from the $i$-th stage $\bm{F}_i \in \mathbb{R}^{H_i \times W_i \times B_i}$, with $i \in \{1, 2,3,4\}$, we first project them into a low-dimensional space $C$, obtaining $\bm{F}'_i \in \mathbb{R}^{H_i \times W_i \times C}$. Then, we compute the context representation $\boldsymbol{\Omega}_i \in \mathbb{R}^{1 \times C}$ as the projection of the globally pooled visual features:
\begin{equation}
    \boldsymbol{\Omega}_i = \texttt{MLP}(\texttt{GAP}(\bm{F}'_i)),
\end{equation}
where \texttt{MLP} denotes a two-layer network made of $1\times1$ convolutions and $\texttt{GAP}$ a global average pooling operation.
Finally, we obtain the relevance of each channel by passing $\boldsymbol{\Omega}_i$ through a sigmoid function $\sigma$ and compute the contextualized features $\bm{F}^c_i \in \mathbb{R}^{H_i \times W_i \times C}$ by:
\begin{equation}
    \label{eq:se}
    \bm{F}^c_i = \bm{F}'_i \odot \sigma(\boldsymbol{\Omega}_i) + \bm{F}'_i,
\end{equation}
where $\odot$ represents the Hadamard (element-wise) product between $\boldsymbol{\Omega}_i$ and all the pixels in $\bm{F}_i$. 
Note that all these operations, employed to restore deformable attention characteristics, are highly efficient, being composed only by $1\times1$ convolutions, normalizations, or element-wise products.

\myparagraph{Feature Aggregation}
Having obtained the contextualized features, we now aggregate them to construct the features pyramid network. 
To do so, we follow previous efficient segmentation works \cite{yoso, bolya2019yolact} relying on deformable convolutions \cite{deformable_conv} to fuse features coming from different scales.

In practice, given the features $\bm{F}^c_i$, with $i \in \{1,2,3,4\}$, where $i=4$ corresponds to the lowest resolution, we compute the intermediate FPN features $\widehat{\bm{F}}_i$ as follows:
\begin{equation}
\widehat{\bm{F}}_i =
\begin{cases}
    \texttt{DefConv}(\bm{F}^c_{i} + \texttt{Proj}(\texttt{GAP}(\bm{F}_{i}))), \; i=4 \\
    \texttt{DefConv}(\bm{F}^c_{i} + \texttt{Up}(\widehat{\bm{F}}_{i+1})), \; i=2,3 \\
    \bm{F}^c_{i} + \texttt{Up}(\widehat{\bm{F}}_{i+1}), \; i=1
\end{cases}
\end{equation}
where \texttt{Proj} indicates a linear projection, \texttt{DefConv} is the deformable convolution and \texttt{Up} is an upsampling operation. Note that the $\widehat{\bm{F}}_4$ is obtained starting using a scene representation that further injects the global context into the FPN, while for the others we mix the upsampled intermediate features of the FPN with the higher-resolution features.

As illustrated in \cref{fig:architecture}, the three lowest resolution features $\widehat{\bm{F}}_i$ ($i \in \{2,3,4\}$) are employed as visual features in the PEM-CA (see \cref{sec:pemca}), while the highest resolution feature $\widehat{\bm{F}}_1$ is used for computing the final predictions.

\section{Experiments}
\label{exp}

\subsection{Experimental Protocol}
\myparagraph{Datasets} 
We evaluate our method on two segmentation datasets: Cityscapes \cite{cityscapes} and ADE20K \cite{ade}.
The Cityscapes dataset features $19$ distinct classes situated in an urban environment, further classified into $8$ \textit{things} and $11$ \textit{stuff} categories. On the other hand, ADE20K is a comprehensive dataset consisting of $150$ diverse classes, encompassing $100$ \textit{things} and $50$ \textit{stuff} categories.

\myparagraph{Baselines}
We conduct a comprehensive comparison of \architectureSpace with both state-of-the-art task-specific and multi-task architectures for panoptic and semantic segmentation. 

\myparagraph{Metrics} 
For semantic segmentation, we employ mean Intersection over Union (mIoU) \cite{mIoU} to measure the performance. For panoptic segmentation, we rely on the Panoptic Quality metric (PQ) \cite{pq}, which encapsulates the overall performance of the models. PQ is defined as the product of two components: Segmentation Quality, which takes into account the Intersection over Union (IoU) between correctly classified segments, and Recognition Quality, which scores the classification accuracy. Furthermore, we report the PQ averaged only on \textit{thing} classes (PQ$_{th}$) and on \textit{stuff} classes (PQ$_{st}$).
If not stated otherwise, latency and FPS are measured using PyTorch 1.12 in FP32 on a V100 GPU with a batch size of 1, using a standard resolution 2048$\times$1024 on Cityscapes and by taking the average runtime on the entire validation set on ADE20K.

\subsection{Implementation Details}
We train our models with AdamW \cite{adamw} optimizer paired with a Cosine learning rate schedule \cite{cosine}. Specifically, we set a learning rate of $0.0007$ for Cityscapes and $0.0004$ for ADE20K, with a $0.1$ multiplier on the backbone. We adopt a batch size of $32$ and the weight decay is set to $0.05$ for both datasets. The models are trained for 90k iterations on Cityscapes and 160k iterations on ADE20K.


\myparagraph{Losses} In terms of loss functions, we follow the configuration established in Mask2Former \cite{m2f}. The classification head is supervised using Binary Cross-entropy (BCE), while for the masks, a combination of BCE and Dice loss \cite{dice} is employed. To balance the influence of these distinct loss components, weight factors of 2.0, 5.0, and 5.0 are assigned to BCE for classification, BCE for masks, and Dice loss, respectively. Furthermore, deep supervision is enabled at each transformer decoder block by default. 

\myparagraph{Architecture}
Unless explicitly specified, our models employ a ResNet50 \cite{resnet} pretrained on ImageNet-1k \cite{imagenet} as backbone. The transformer decoder utilized in our architecture is derived from Mask2Former \cite{m2f}, where the masked cross-attention layers are replaced by our proposed PEM-CA. The architectural configuration consists of two transformer decoder stages with a hidden dimension of $256$ and $8$ heads in the attention layers. An expansion factor of $8$ is applied in the feed-forward networks and a fixed number of $100$ object queries is employed. Furthermore, the hidden dimension of our pixel decoder is set to $128$.

\subsection{Results}

\begin{table}[t]
\setlength{\tabcolsep}{2pt}
\centering
\begin{tabular}{@{}l|ccc|crr@{}}
\toprule
Method                            & PQ        & PQ$_{th}$ & PQ$_{st}$ & FPS            & FLOPs   & Params    \\ \midrule
Mask2Former \cite{m2f}            & 62.1      & -         & -         & 4.1            & 519G    & 44.0M     \\
UPSNet \cite{upsnet}              & 59.3      & 54.6      & 62.7      & 7.5            &  -      &  -        \\
LPSNet \cite{lpsnet}              & 59.7      & 54.0      & 63.9      & 7.7            &   -     &  -        \\
PanDeepLab \cite{pandeeplab} & 59.7      & -         & -         & 8.5            &  -      &  -        \\
FPSNet \cite{fpsnet}              & 55.1      & -         & -         & 8.8$^\dagger$  &  -      &  -        \\
RealTimePan \cite{realtimepan}    & 58.8      & 52.1      & 63.7      & 10.1           &  -      & -         \\
YOSO \cite{yoso}                  & 59.7      & 51.0      & 66.1      & 11.1           & 265G    & 42.6M     \\  \midrule
\textbf{\architecture}            & 61.1 & 54.3 & 66.1 & 13.8 & 237G  & 35.6M     \\ \bottomrule
\end{tabular}
\caption{\textbf{Panoptic segmentation on Cityscapes with 19 categories.} $\dagger$: measured on a Titan GPU. ResNet50 is employed as backbone for all the architectures.}
\label{tab:cit_pan}
\end{table}
\begin{table}[t]
\centering
\resizebox{\columnwidth}{!}
{
\setlength{\tabcolsep}{3pt}
{\begin{tabular}{@{}l|ccc|ccc@{}}
\toprule
Method                        & PQ   & PQ$_{th}$ & PQ$_{st}$ & FPS   & FLOPs & Params \\ \midrule
BGRNet \cite{bgrnet}          & 31.8 & 34.1      & 27.3      & -     & -    & - \\
MaskFormer \cite{maskformer}  & 34.7 & 32.2      & 39.7      & 29.7  & 86G  & 45.0M    \\
Mask2Former \cite{m2f}        & 39.7 & 39.0      & 40.9      & 19.5  & 103G & 44.0M \\
kMaxDeepLab \cite{kmax}       & 42.3 & -         & -         & -     & -    & - \\
YOSO \cite{yoso}              & 38.0 & 37.3      & 39.4      & 35.4  & 52G  & 42.0M \\ \midrule
\textbf{\architecture}        & 38.5  & 37.0 & 41.1      &  35.7 & 47G & 35.6M   \\ \bottomrule
\end{tabular}}}
\caption{\textbf{Panoptic segmentation on ADE20k with 150 categories.} ResNet50 is employed as backbone for all the architectures. FLOPs are measured at resolution 640x640.} \vspace{-11pt}
\label{tab:ade_pan}
\end{table}

\myparagraph{Panoptic Segmentation on Cityscapes}  
\cref{tab:cit_pan} presents the results in the panoptic setting for Cityscapes. The key challenge of this dataset is to reach reasonable FPS due to the high-resolution of input images, stressing the need for efficient approaches. Notably, \architectureSpace demonstrates remarkable performance, achieving 61.1 PQ, while being the fastest architecture, with 13.8 FPS. PEM outperforms all competitor models except for the heavyweight and slower Mask2Former. Indeed, for a loss of 1 PQ, PEM is twice as fast as Mask2Former. Furthermore, our architecture exhibits a substantial improvement, with a 1.4 PQ gain and 2.7 FPS advantage over the second-fastest approach, YOSO. Remarkably, while showing similar PQ$_{st}$ compared to YOSO, our architecture outperforms it on PQ$_{th}$ (+3.3). Overall, PEM exhibits the most favorable performance-speed trade-off among all the models.

\myparagraph{Panoptic Segmentation on ADE20K} 
The ADE20K panoptic results are outlined in \cref{tab:ade_pan}. The large-scale dimension of the dataset together with the high number of classes make ADE20K one of the most challenging benchmarks in segmentation. Nonetheless, \architectureSpace attains a PQ of 38.5 at 35.7 FPS. 
Within this context, only heavy and slow architectures, such as Mask2Former and kMaxDeepLab, manage to surpass the PQ of PEM, while being slower. In particular, Mask2Former is 16.2 FPS slower than PEM. When compared to the fastest competitor, YOSO, our model demonstrates an improvement of 0.5 PQ. In essence, \architectureSpace demonstrates the best performance-speed trade-off, confirming the findings observed on the Cityscapes dataset.

\begin{table}[t]
\vspace{-0.3cm}
\setlength{\tabcolsep}{1pt}
\centering
\begin{tabular}{@{}l|c|c|crr@{}}
\toprule
Method                       & Backbone        &  mIoU                & FPS                  & FLOPs             & Params     \\ \midrule
\multicolumn{6}{c}{Task-specific Architectures} \\ \midrule
BiSeNetV1 \cite{bisenet}     & R18           & 74.8                 & 65.5$^\dagger$       & 55G               & 49.0M    \\
BiSeNetV2-L\cite{bisenetv2}    & -               & 75.8                 & 47.3$^\ddagger$      & 119G              & -        \\
STDC1-Seg75 \cite{stdc}      & STDC1           & 74.5                 & 74.8$^\dagger$       & -                    & -       \\
STDC2-Seg75 \cite{stdc}      & STDC2           & 77.0                 & 58.2$^\dagger$       & -                    & -       \\
DDRNet-23-S \cite{ddrnet}    & -               & 77.8     & 108.1              & 36G      & 5.7M     \\
DDRNet-23 \cite{ddrnet}      & -               & 79.5     & 51.4               & 143G     & 20.1M    \\
PIDNet-S \cite{pidnet}       & -               & 78.8     & 93.2               & 46G      & 7.6M     \\
PIDNet-M \cite{pidnet}       & -               & 80.1     & 39.8               & 197G     & 34.4M    \\
PIDNet-L \cite{pidnet}       & -               & 80.9     & 31.1               & 276G     & 36.9M    \\ \midrule
\multicolumn{6}{c}{Multi-task Architectures} \\ \midrule
MaskFormer \cite{maskformer}       & R101      & 78.5     & 10.1               & 559G     & 60.2M       \\
Mask2Former \cite{m2f}       & R50             & 79.4     & 6.6                & 523G     & 44.0M        \\
YOSO \cite{yoso}             & R50             & 79.4     & 11.1               & 268G     & 42.6M        \\  \midrule
\textbf{\architecture}       & STDC1           & 78.3   & 24.3    & 92G     & 17.0M    \\
\textbf{\architecture}       & STDC2           & 79.0   & 22.0    & 118G    & 21.0M    \\
\textbf{\architecture}       & R50             & 79.9   & 13.8    & 240G    & 35.6M    \\ \bottomrule
\end{tabular}
\caption{\textbf{Semantic segmentation on Cityscapes with 19 categories.} $\dagger$: resolution of 1536x768. $\ddagger$: resolution of 1024x512.
\vspace{-10pt}}
\label{tab:cit_sem}
\end{table}

\myparagraph{Semantic Segmentation on Cityscapes} 
\cref{tab:cit_sem} presents the results for Cityscapes in the semantic setting. PEM achieves a mIoU of 79.9 at 13.8 FPS. Although slower, PEM shows superior performance \wrt all competitors with the exception of the highly engineered and tailored PIDNet. Conversely, when compared to multi-task architectures, PEM obtains a gain of 0.5 mIoU over both Mask2Former and YOSO and 1.4 mIoU over MaskFormer, being faster than them by more than 2 FPS. In summary, PEM emerges as the general architecture with the best trade-off between performance and latency for semantic segmentation on Cityscapes. Additionally, when equipped with STDC\cite{stdc} as backbone, PEM maintains its performance while demonstrating a consistent improvement in speed.

\begin{table}[t]
\vspace{-0.3cm}
\setlength{\tabcolsep}{1.7pt}
\centering
\begin{tabular}{@{}l|c|c|crr@{}}
\toprule
Method        & Backbone &  mIoU  & FPS     & FLOPs    & Params \\ \midrule
\multicolumn{6}{c}{Task-specific Architectures}                 \\ \midrule
BiSeNetV1 \cite{bisenet}      & R18    &  35.1      &  143.1      &  15G   & 13.3M       \\
BiSeNetV2-L \cite{bisenetv2}  & -        &  28.5      &  106.7      &  12G   & 3.5M       \\
STDC1 \cite{stdc}             & STDC1    &  37.4      &  116.1      &  8G    & 8.3M       \\
STDC2 \cite{stdc}             & STDC2    &  39.6      &  78.5       &  11G   & 12.3M       \\
DDRNet-23-S \cite{ddrnet}     & -        &  36.3      & 96.2        & 4G     & 5.8M       \\
DDRNet-23 \cite{ddrnet}       & -        &  39.6      & 94.6        & 18G    & 20.3M      \\
PIDNet-S \cite{pidnet}        & -        &  34.8      & 73.5        & 6G     & 7.8M       \\
PIDNet-M \cite{pidnet}        & -        &  38.8      & 73.3        & 22G    & 28.8M       \\
PIDNet-L \cite{pidnet}        & -        &  40.5      & 65.4        & 34G    & 37.4M       \\ \midrule
\multicolumn{6}{c}{Multi-task Architectures}                     \\ \midrule
MaskFormer \cite{maskformer}  & R50      & 44.5   & 29.7        &  55.1G        & 41.3M       \\ 
Mask2Former \cite{m2f}        & R50      & 47.2   & 21.5        & 70.1G         &  44.0M      \\
YOSO \cite{yoso}              & R50      & 44.7   & 35.3       &  37.3G        &  42.0M      \\ \midrule
\textbf{\architecture}        & STDC1  & 39.6   &  43.6   & 16.0G   & 17.0M    \\
\textbf{\architecture}        & STDC2  & 45.0   &  36.3       &  19.3G   &  21.0M   \\ 
\textbf{\architecture}        & R50    & 45.5   &  35.7   &  46.9G   &  35.6M  \\\bottomrule
\end{tabular}
\caption{\textbf{Semantic segmentation on ADE20K with 150 categories.} FLOPs are measured at resolution 512x512.}\vspace{-10pt}
\label{tab:ade_sem}
\end{table}

\vspace{-5pt}
\myparagraph{Semantic Segmentation on ADE20K} 
\cref{tab:ade_sem} outlines the results for semantic segmentation on the large-scale ADE20K dataset. Although it features over 150 distinct classes, \architectureSpace attains a noteworthy 45.5 mIoU at 35.7 FPS. When compared to general-purpose architectures, PEM shows an outstanding performance vs efficiency trade-off, with only Mask2Former able to achieve higher mIoU, while being almost 15 FPS slower. 
Contrary to the Cityscapes results, PEM outperforms all task-specific approaches by a large margin, showcasing a 5 mIoU gain over the highest-performing competitor, PIDNet-L. Furthermore, also in this scenario, with the integration of an efficient backbone, such as STDC1 or STDC2, PEM yields remarkable performance while significantly reducing inference time.

\subsection{Ablation Study}

We perform ablation studies on the various components of the proposed architecture, in order to assess the contribution of each module. All the ablations are performed on Cityscapes \cite{cityscapes} in the panoptic setting.

\myparagraph{\method} 
\cref{tab:efficient_CA} serves as a comprehensive comparison for evaluating the efficacy of \methodSpace against traditional cross-attention mechanisms, shedding light on the impact of removing masking and prototype selection from our module. The absence of the masking mechanism, designed to enhance focus on foreground regions, results in a noticeable 3.3 panoptic quality (PQ) loss. This highlights the crucial role played by the masking mechanism in guiding the model attention to relevant regions, especially those representing foreground objects. Complete removal of the prototype selection strategy, with token aggregation by summation, akin to SwiftFormer \cite{swift}, leads to a more consistent performance loss of 13.4 PQ and almost 20 PQ$_{th}$. The substantial performance drop underlines the critical point that tokens cannot be naively collapsed. This emphasizes the necessity of a careful selection strategy, demonstrating the importance of our proposed prototype selection mechanism for optimal token representation. Notably, all observed performance drops are mostly associated with the \textit{things} category. This suggests that the prototype selection strategy is particularly critical for distinguishing between different instances of objects. 
Overall, \methodSpace demonstrates superior performance compared not only to standard cross-attention but also its masked counterpart \cite{m2f}, while being significantly faster than both methods. The achieved PQ$_{th}$ improvement exceeds 2, showcasing the efficacy of the proposed \methodSpace in the realm of panoptic segmentation.

\begin{table}[t]
\setlength{\tabcolsep}{3pt}
\centering
\begin{tabular}{@{}l|ccc|cc@{}}
\toprule
Method                                        & PQ   & PQ$_{th}$ & PQ$_{st}$ & Latency  & FLOPs    \\  \midrule
CA \cite{maskformer}                          & 58.4 &   47.4    &   66.4    & 11.4 ms  &  228G    \\ 
Masked CA \cite{m2f}                          & 60.4 &   52.0    &   66.5    & 17.4 ms  &  246G    \\  \midrule
\makecell[l]{PEM-CA \\ w/o Prototypes}        & 48.7 &   24.7    &   66.1    & 5.6 ms   &  218G    \\  \specialrule{0.05mm}{0.5pt}{1pt}    
\makecell[l]{PEM-CA \\ w/o Masking}           & 57.8 &   47.0    &   65.6    & 6.8 ms   &  221G    \\  \specialrule{0.05mm}{0.5pt}{2.5pt}
\textbf{PEM-CA}                               & \textbf{61.1} & \textbf{54.3} & \textbf{66.1} & 9.8 ms &  237G   \\ \bottomrule
\end{tabular} 
\caption{\textbf{Ablation of PEM-CA on Cityscapes}. We report the cumulative latency of the cross-attention modules.} \vspace{-2pt}
\label{tab:efficient_CA}
\end{table}

\myparagraph{Lightweight Pixel Decoder} 
\cref{tab:pixel_dec} illustrates the behavior of \architectureSpace when certain components are excluded from the lightweight pixel decoder and it presents a comparison with MaskFormer and Mask2Former pixel decoders. Specifically, the absence of CSM modules results in more than 1 PQ and almost 3 PQ$_{th}$ loss, thereby demonstrating the need of global context modeling and dynamic weights. Furthermore, substituting Deformable convolutions with standard ones exacerbates the PQ loss, yielding a decrease of 4 PQ and nearly 7 PQ$_{th}$. This underscores the critical role of kernel deformability, especially for discerning multiple instances within the \textit{things} category. When compared to the heavy convolutional decoder of MaskFormer, our decoder demonstrates superior performance with a higher PQ (+3.3) and higher FPS (+2.0). This attests to the effectiveness of our approach, which integrates the strengths of the Mask2Former decoder in a convolutional manner. Notably, while our decoder and the Mask2Former decoder achieve similar performance levels, our implementation stands out for its efficiency, running at twice the FPS.

\begin{table}[t]
\setlength{\tabcolsep}{1pt}
\centering
\begin{tabular}{@{}l|ccc|cc@{}}
\toprule
Method                               & PQ   & PQ$_{th}$ & PQ$_{st}$ & Latency  & FLOPs  \\ \midrule
PEM w/ MF decoder \cite{maskformer} & 57.8 &   48.5    &  64.5     & 84.7ms  & 414G    \\
PEM w/ M2F decoder \cite{m2f}       & 61.4 &   54.6    &  66.3     & 131.6ms & 497G   \\ \midrule
PEM w/o CSM                    & 60.0 &   51.5    &  66.1     & 71.9ms  & 237G    \\
PEM w/o deformable                   & 57.1 &   47.6    &  64.0     & 69.9ms  & 250G     \\ \midrule
\textbf{\architecture}       & \textbf{61.1} &   \textbf{54.3}    &  \textbf{66.1}     & 72.4ms  &  237G  \\ \bottomrule
\end{tabular}
\caption{\textbf{Ablation of Pixel Decoders on Cityscapes.} We report the total latency for the whole model.}\vspace{-0.5cm}
\label{tab:pixel_dec}
\end{table}

\begin{figure}[t]
    \centering
    \includegraphics[width=\linewidth]{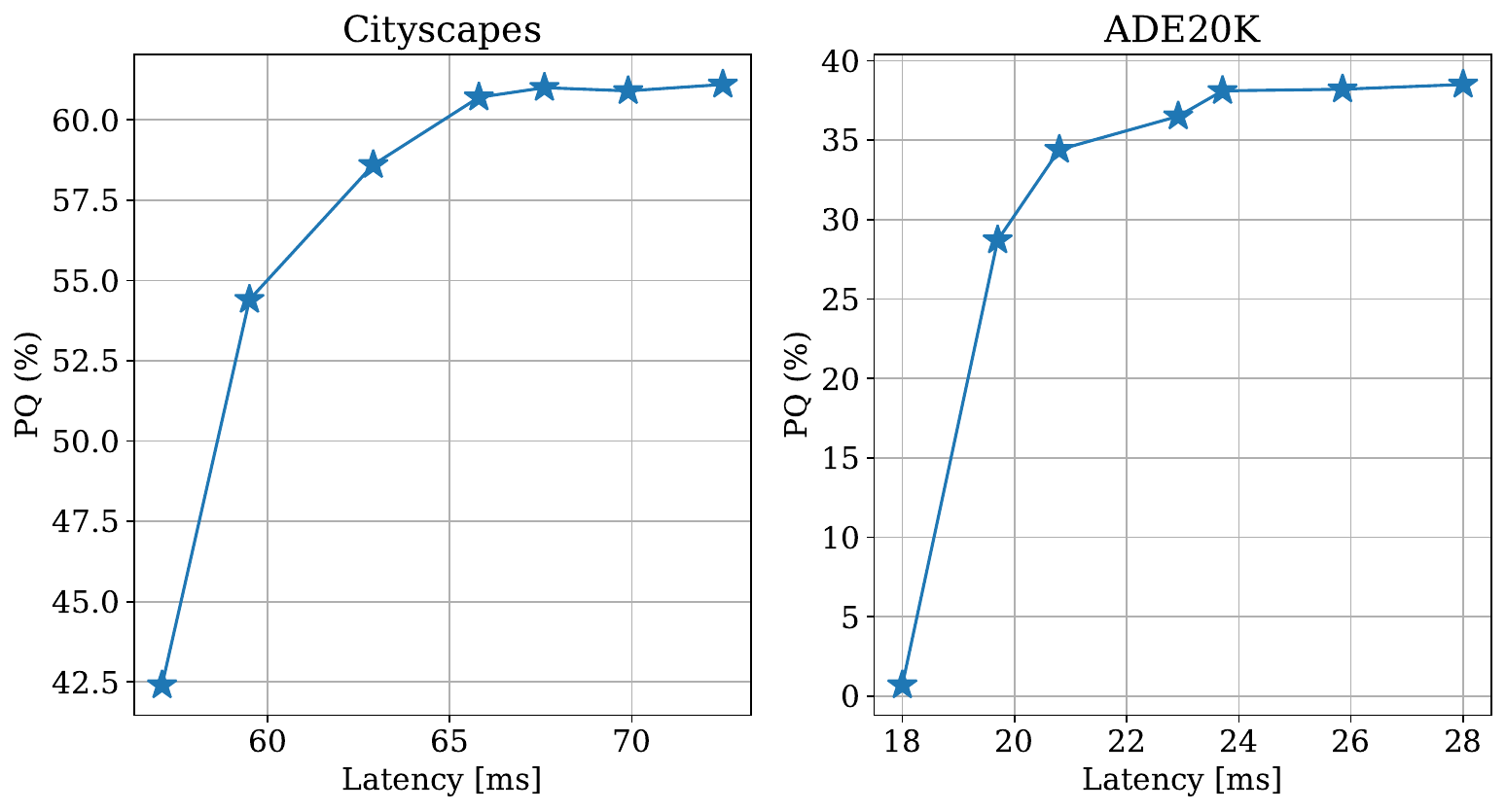}
    \caption{\textbf{PQ versus latency on Cityscapes and ADE20K.} We report performance and latency across different numbers of PEM transformer decoder blocks, ranging from zero to six.} \vspace{-10pt}
    \label{fig:num_dec_layers}
\end{figure}

\myparagraph{Number of Transformer Decoder Layers} 
Figure \ref{fig:num_dec_layers} shows the variation in performance and latency when the number of transformer decoding layers varies from zero, where prediction is carried out solely on initialized queries, to six, representing the complete configuration. Notably, PEM exhibits robust performance even with a single stage of transformer decoder, \ie three decoding layers, offering a flexible trade-off between performance and speed. Moreover, on the challenging ADE20K dataset, the model attains nearly zero PQ without cross-attention layers, emphasizing the need for an efficient query refinement process.

\section{Conclusions}
\label{conclusions}
Our work introduces \architecture, a transformer-based architecture addressing the efficiency challenges posed by current models in image segmentation tasks. Leveraging a novel prototype-based cross-attention mechanism and an efficient multi-scale feature pyramid network, PEM achieves outstanding performance in both semantic and panoptic segmentation tasks on multiple datasets. PEM's efficiency surpasses state-of-the-art multi-task architectures and competes favorably with computationally expensive baselines, demonstrating its potential for deployment in resource-constrained environments. This work represents a significant step towards efficient and high-performance segmentation solutions using transformer-based architectures.
\paragraph{Acknowledgements}
This study was carried out within the FAIR - Future Artificial Intelligence Research and received funding from the European Union Next-GenerationEU (PIANO NAZIONALE DI RIPRESA E RESILIENZA (PNRR) – MISSIONE 4 COMPONENTE 2, INVESTIMENTO 1.3 – D.D. 1555 11/10/2022, PE00000013). This manuscript reflects only the authors’ views and opinions, neither the European Union nor the European Commission can be considered responsible for them. We acknowledge the CINECA award under the ISCRA initiative, for the availability of high performance computing resources and support.

{
    \small
    \bibliographystyle{ieeenat_fullname}
    \bibliography{main}
}

\clearpage

\setcounter{page}{1}
\maketitlesupplementary

\section{Additional Implementation Details}

In the subsequent section, we offer supplementary information about the dataset utilized in our experiment, indicating some additional training parameters used for each respective dataset.

\myparagraph{Cityscapes}
Cityscapes is a dataset containing high-resolution images ($1024 \times 2048$ pixels) that depict urban street views from an egocentric perspective. The dataset comprises 2975 training images, 500 validation images, and 1525 testing images, encompassing 19 distinct classes.

For both segmentation tasks, we use a fixed crop size of $512 \times 1024$ during training. During inference, the full-resolution image is used.

\myparagraph{ADE20K}
The ADE20K dataset contains 20,000 images for training and 2,000 images for validation, captured at diverse locations and featuring a wide range of objects. The images vary in size.

Following the methodology proposed in \cite{m2f}, a unique crop size is utilized for each segmentation task during the training process. Semantic segmentation uses a fixed crop size of $512 \times 512$, while panoptic segmentation uses a fixed crop size of $640 \times 640$. During inference, the shorter side of the image is resized to fit the corresponding crop size.

\myparagraph{Metrics} The metric adopted for semantic segmentation, mean Intersection over Union, is defined as:

\begin{equation}
   \text{mIoU} =  \frac{1}{K} \sum_{i=1}^K  \left| \frac{\mathcal{Y} \cap Y}{\mathcal{Y} \cup Y} \right|,
   \label{eq:final_pred}
\end{equation}

where K is the number of classes, the numerator is the intersection between the predicted mask $\mathcal{Y}$ and the ground truth $Y$, while the denominator is their union.

Considering panoptic segmentation, Panoptic Quality takes into account both the quality of object segmentation and the correctness of assigning semantic labels to the segmented objects. Formally, it is defined as follows:
\begin{equation}
    \mathrm{PQ} = \frac{\sum_{i \in TP} \mathrm{IoU}}{|TP| + \frac{1}{2}|FP| + \frac{1}{2} |FN|}.
\end{equation}

In practice, it can be seen as the product of Segmentation Quality (SQ) and Recognition Quality (RQ):
\begin{equation}
    \mathrm{PQ}=\underbrace{\frac{\sum_{i \in TP} \operatorname{IoU}}{|TP|}}_{\text{SQ}} \times \underbrace{\frac{|TP|}{|TP|+\frac{1}{2}|FP|+\frac{1}{2}|FN|}}_{\text{RQ}}.
\end{equation}
SQ measures the similarity between correctly predicted segments and their corresponding ground truths while PQ measures the overall ability of the model in identifying and classifying objects or segments.

\myparagraph{Baselines}
In \cref{tab:ade_sem}, the task-specific architectures have been retrained from scratch given that these models were not tested on the ADE20K dataset. To ensure fairness in comparison with our model, we performed hyper-parameter tuning for the different architectures. Nonetheless, our comprehensive evaluation has revealed that our pipeline demonstrates suitability across diverse approaches, ultimately yielding the highest results. Specifically, we trained these models for 160,000 iterations using AdamW \cite{adamw} optimizer, Cosine learning rate scheduler \cite{cosine}, initial learning rate set to 0.0004, and weight decay set to 0.05. The crop size remained consistent with our experiment, at $512 \times 512$.

\section{Additional Ablation Study}

\myparagraph{Number of queries} 
The results for Cityscapes varying number of queries are reported in the \cref{tab:abl_numqueries}. 
Using 50 queries leads to a performance drop while increasing them to 200 is not beneficial while causing a substantial increase in complexity.

\begin{table}[t]
    \centering
    \begin{tabular}{@{}c|ccc|c@{}} 
    \toprule
     N & PQ   & PQ$_{th}$ & PQ$_{st}$ & FLOPs \\ 
    \midrule
    50  & 58.7 & 48.8 & 65.9 & 225G \\
    100 & \textbf{61.1} & \textbf{54.3} & 66.1 & 237G \\
    200 & 61.0 & 53.7 & \textbf{66.4} & 275G \\
    \bottomrule
    \end{tabular}
    \caption{Ablation on number of queries on Cityscapes.} \vspace{-3mm}
    \label{tab:abl_numqueries}
\end{table}

\section{Qualitative Results}
\label{qualitative}

We showcase qualitative results of \architectureSpace on the Cityscapes and ADE20K datasets, highlighting its performance both in semantic and in panoptic segmentation. Our evaluations involve comparisons with resource-intensive architectures like Mask2Former \cite{m2f} and lightweight alternatives such as YOSO \cite{yoso}. PEM exhibits comparable performance to Mask2Former while demonstrating superiority over YOSO. Specifically, our model excels in distinguishing different instances in the panoptic setting and displays a lower number of false positives.

\begin{figure*}[t]
    \centering 
    \includegraphics[width=1\linewidth]{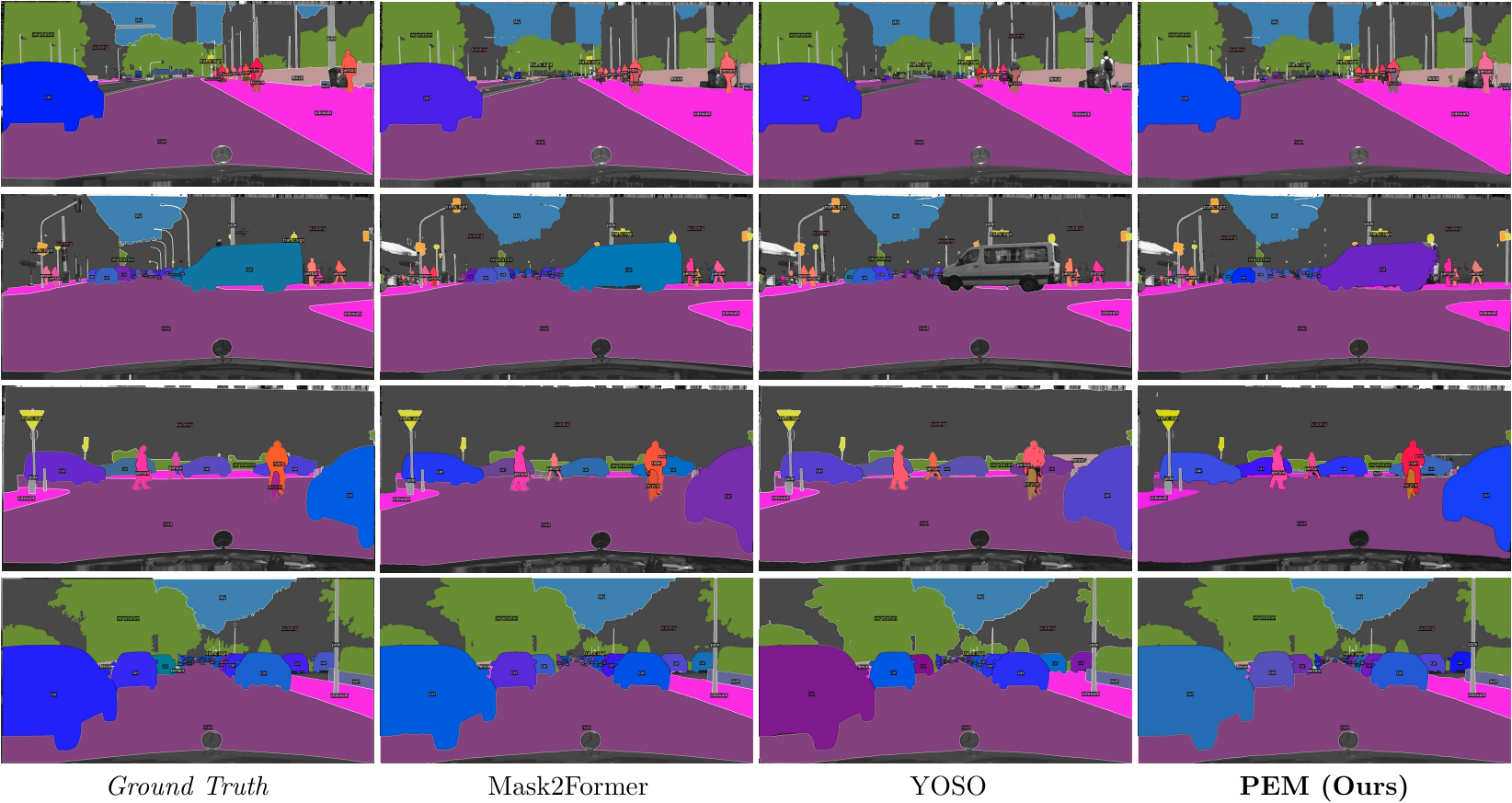}
    \caption{\textbf{Qualitative results} of PEM \textit{v.s.} Mask2Former \cite{m2f} and YOSO \cite{yoso} on \textit{panoptic segmentation} on Cityscapes.} \vspace{-6pt} 
    \label{fig:qual_cit_pan}
\end{figure*}

\begin{figure*}[t]
    \centering 
    \includegraphics[width=1\linewidth]{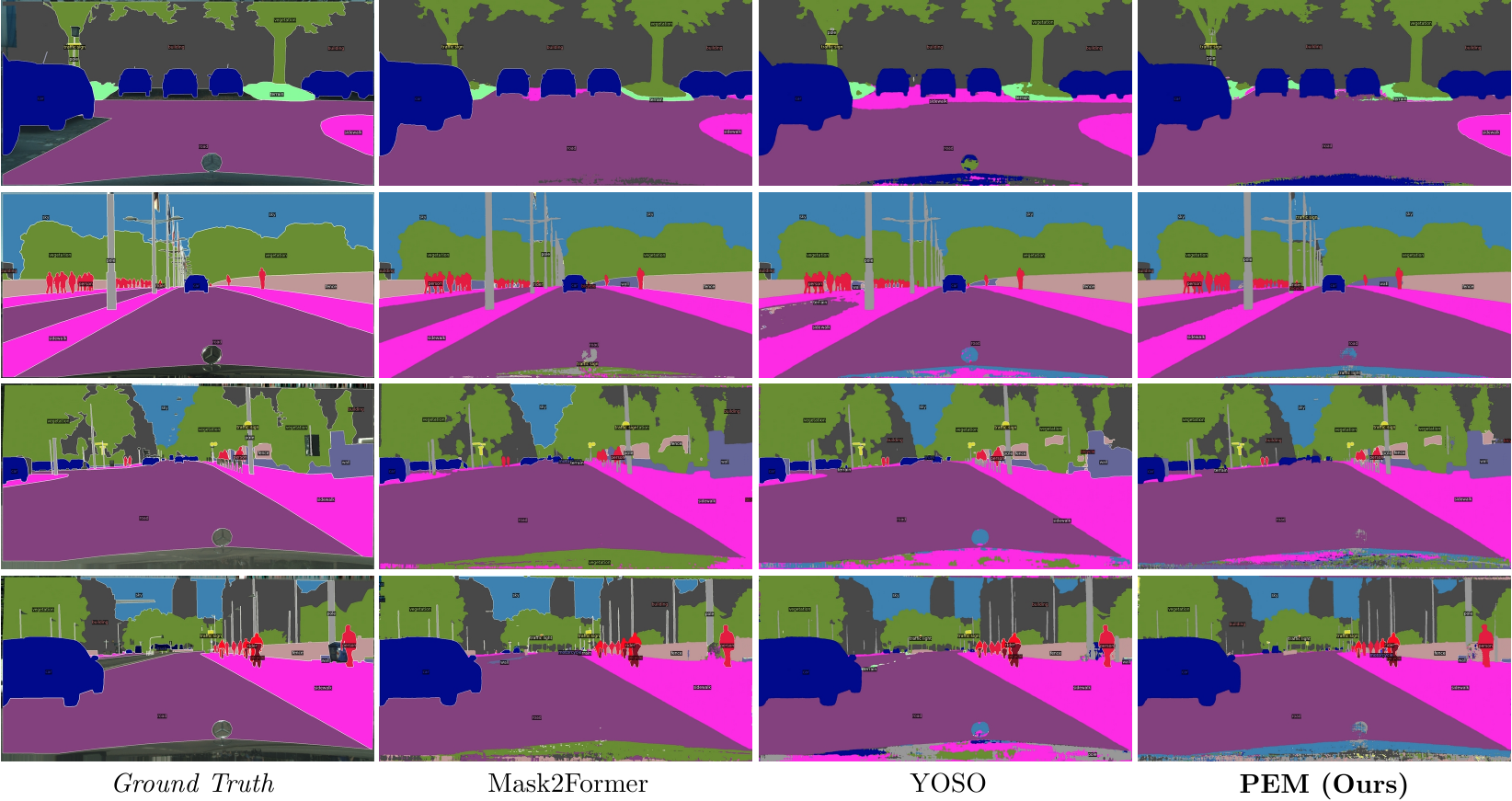}
    \caption{\textbf{Qualitative results} of PEM \textit{v.s.} Mask2Former \cite{m2f} and YOSO \cite{yoso} on \textit{semantic segmentation} on Cityscapes.} \vspace{-6pt} 
    \label{fig:qual_cit_sem}
\end{figure*}

\begin{figure*}[t]
    \centering 
    \includegraphics[width=0.9\linewidth]{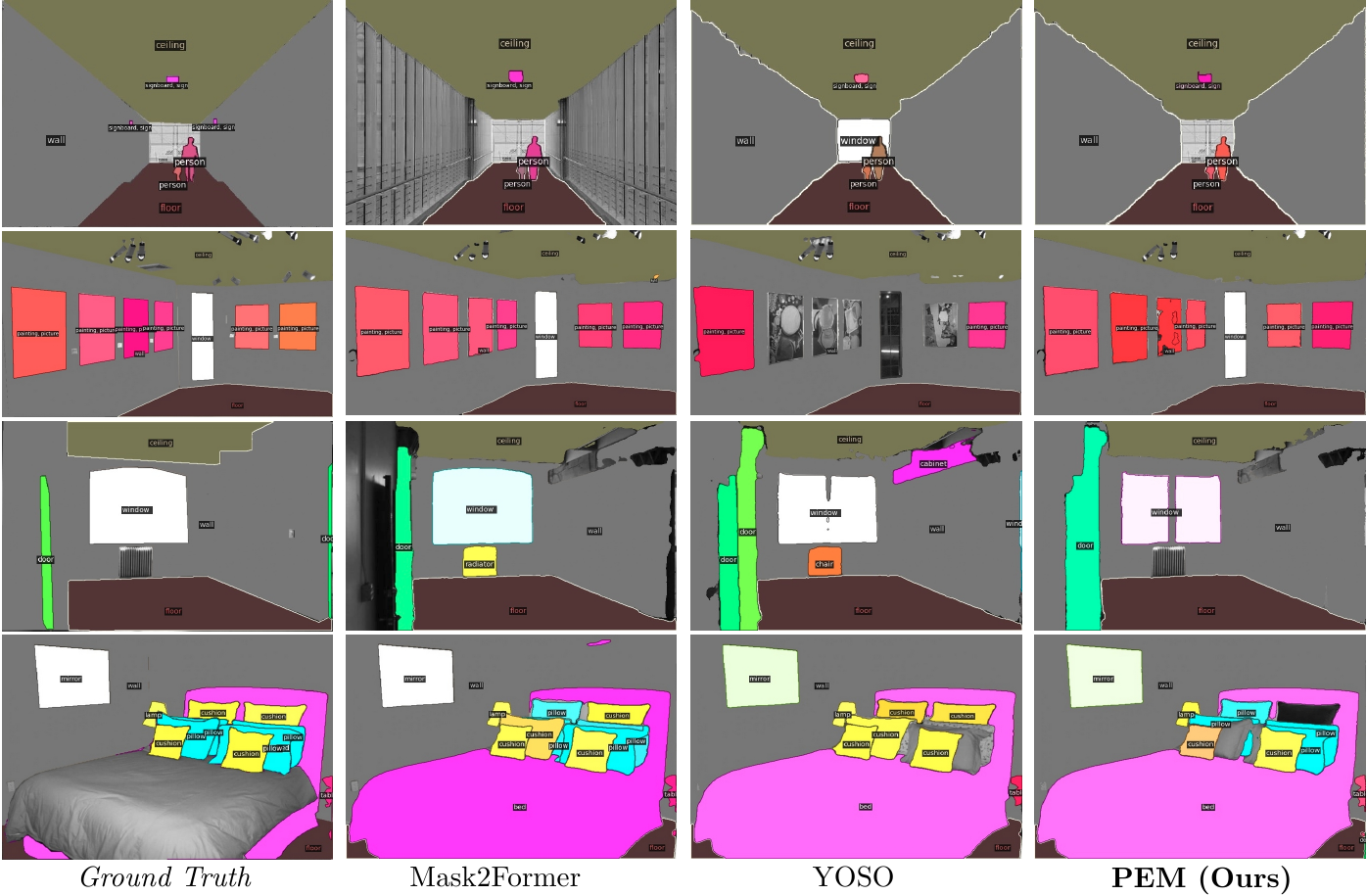}
    \vspace{-4pt} 
    \caption{\textbf{Qualitative results} of PEM \textit{v.s.} Mask2Former \cite{m2f} and YOSO \cite{yoso} on \textit{panoptic segmentation} on ADE20K.} \vspace{-8pt} 
    \label{fig:qual_ade_pan}
\end{figure*}

\vspace{-15pt} 

\begin{figure*}[t]
    \centering 
    \includegraphics[width=0.9\linewidth]{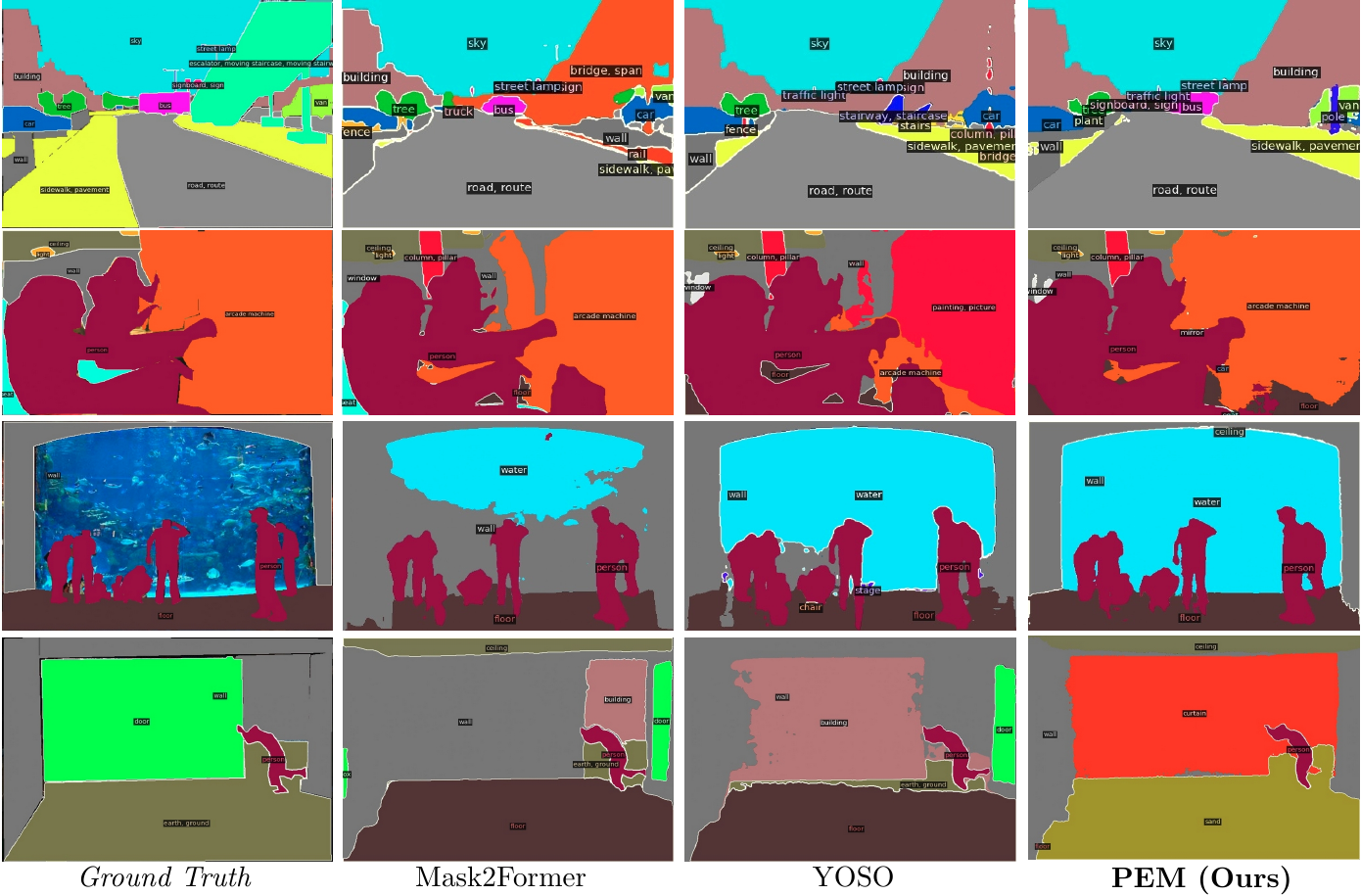}
    \vspace{-4pt} 
    \caption{\textbf{Qualitative results} of PEM \textit{v.s.} Mask2Former \cite{m2f} and YOSO \cite{yoso} on \textit{semantic segmentation} on ADE20K.} \vspace{-8pt} 
    \label{fig:qual_ade_sem}
\end{figure*}


\end{document}